\documentclass[a4paper,twoside]{article}
\usepackage{cite}

\usepackage{epsfig}
\usepackage{subcaption}
\usepackage{calc}
\usepackage{amssymb}
\usepackage{amstext}
\usepackage{amsmath}
\usepackage{amsthm}
\usepackage{multicol}
\usepackage{pslatex}
\usepackage{apalike}
\usepackage{algorithm2e}
\usepackage{hyperref}
\hypersetup{
    colorlinks=true, % Enable colored links (true removes boxes)
    linkcolor=black, % Color of internal links (sections, pages, etc.)
    urlcolor=black, % Color of external links (URLs)
    citecolor=black % Color of citations
}

\usepackage[symbol,bottom]{footmisc}

\usepackage{SCITEPRESS}     % Please add other packages that you may need BEFORE the SCITEPRESS.sty package.

\begin{document}

\title{Automated Detection of Defects on Metal Surfaces using Vision Transformers}

\author{\authorname{Toqa Alaa\sup{1}, Mostafa Kotb\sup{1}, Arwa Zakaria\sup{1}, Mariam Diab\sup{1}, and Walid Gomaa\sup{1,2}}
\affiliation{\sup{1} Department of Computer Science and Engineering, Egypt-Japan University of Science and Technology, Alexandria, Egypt.}
\affiliation{\sup{2}Faculty of Engineering, Alexandria University, Alexandria, Egypt.}
\email{\{toqa.alaa, mostafa.kotb, arwa.zakaria, mariam.diab, walid.gomaa\}@ejust.edu.eg}
}


\abstract{Metal manufacturing often results in the production of defective products, leading to operational challenges. Since traditional manual inspection is time-consuming and resource-intensive, automatic solutions are needed. The study utilizes deep learning techniques to develop a model for detecting metal surface defects using Vision Transformers (ViTs). The proposed model focuses on the classification and localization of defects using a ViT for feature extraction. The architecture branches into two paths: classification and localization. The model must approach high classification accuracy while keeping the Mean Square Error (MSE) and Mean Absolute Error (MAE) as low as possible in the localization process. Experimental results show that it can be utilized in the process of automated defects detection, improve operational efficiency, and reduce errors in metal manufacturing.}

\onecolumn \maketitle \normalsize \setcounter{footnote}{0} \vfill

\section{\uppercase{Introduction}}
\label{sec:introduction}

\par 
The manufacturing and reshaping of metal surfaces are critical processes in various industries, including automotive, aerospace, and construction. These processes often result in products with defects such as cracks, dents, scratches, and other surface irregularities. Such defects can compromise the structural integrity and performance of metal products, posing significant challenges to quality control and product usability. Detecting and addressing these defects is crucial to ensure the production of high-quality metal products and to prevent costly operational failures~\cite{wang2021,murakami2019metal,leibfried2006point}.

\par 
Traditionally, defect detection on metal surfaces has relied heavily on manual inspection, where human experts visually examine surfaces for abnormalities. This method is not only time-consuming and labor-intensive but also highly subjective and inconsistent. It is prone to errors and often fails to detect subtle defects that are not easily visible to the human eye. Consequently, there is a compelling need for automated defect detection systems that can accurately and efficiently identify and classify defects on metal surfaces\cite{li2022,Fang2020}.

\par 
Significant progress has been made in developing automated defect detection techniques. Traditional computer vision methods, such as edge detection, thresholding, hough transfrom, and image segmentation, have been extensively explored. These methods typically rely on handcrafted features and rule-based algorithms to identify defects based on predefined criteria. Although these approaches have achieved some success in detecting specific types of defects, they are limited in their ability to handle complex and varied defect patterns. They depend heavily on the expertise of domain-specific engineers and lack the adaptability to new or varied defect types~\cite{metal_image_processing}.

\par 
With the advent of deep learning techniques and the availability of large-scale annotated datasets, there has been a paradigm shift towards employing neural networks for automated defect detection. Convolutional Neural Networks (CNNs) have demonstrated remarkable performance in various computer vision tasks, including image classification, object detection, and semantic segmentation. CNNs can automatically learn discriminative features from raw input data, making them well-suited for defect detection on metal surfaces. Several studies have successfully applied CNNs to detect defects on metal surfaces, achieving high accuracy and demonstrating the potential of deep learning in this domain.

\par 
Despite their promise, CNNs have limitations in defect detection. They typically rely on local receptive fields and hierarchical feature extraction, which may not adequately capture long-range dependencies and global context in images. This limitation is particularly critical when dealing with complex defect patterns that span significant portions of metal surfaces. Additionally, CNNs require large amounts of labeled training data, which can be challenging and time-consuming to acquire for specific defect types or rare occurrences~\cite{Tao2018}.

\par 
To address these limitations, we propose the use of Vision Transformers (ViTs) for automated defect detection on metal surfaces. ViTs, originally introduced for image classification,~\cite{DBLP:journals/corr/abs-2010-11929}
have gained attention for their ability to capture global context and long-range dependencies through self-attention mechanisms~\cite{NIPS2017_3f5ee243} .
This makes them well-suited for capturing intricate defect patterns on metal surfaces. 

\par 
To address the limitations of the current datasets used in metal surface detection, a new dataset called Multi-DET is built. Current datasets don't accurately simulate real-world conditions, as metal surfaces typically have more overlapping and higher number of defects per image. Our new dataset, Mult-DET, addresses these limitations by introducing diversity and increased density per image.

\par 
The primary objectives of this research are twofold: defect classification and defect localization. Defect classification aims to accurately identify the type and nature of each defect. Defect localization aims to precisely predict the boundaries of each defect, facilitating targeted treatment and repair. The model should be able to detect multiple defects in the input image. 

\par
Leveraging the power of pre-trained ViTs on large-scale image datasets like Imagenet, the proposed model utilizes transfer learning to benefit from the learned representations of ViTs, which capture rich visual features. This pre-trained model is able to effectively extract meaningful defect-related features from raw metal surface images. 
So, we propose using Vision Transformers and deep learning techniques to automate defect detection on metal surfaces, aiming to enhance product quality and reduce costly errors in metal manufacturing. 

\par 
The paper is organized as follows. Section \ref{sec:introduction} is an introduction, offering an overview of the problem and the undergoing research. 
Section \ref{sec:related_work} discusses related work. Section \ref{sec:datasets} explores the used datasets and our new dataset.
Section \ref{sec:experimenatl_work} discusses the methodologies used for defect detection and localization. Section \ref{sec:evaluation} gives the experimental work to validate our approach along with analyzing the results, illustrating as well the limitations of our approach. Section \ref{sec:conclusion} summarizes our work and provides an outlook on future directions. 

\section{\uppercase{Related Work}}
\label{sec:related_work}

\par 
Metal defects detection is a critical task in industrial applications. Over the years, researchers have developed methods to identify and classify these defects using machine learning and computer vision ~\cite{wang2021}.
The advancement of these methods increases the efficiency and accuracy of the process to ensure quality and reliability of metal products in industries.

\subsection{Traditional Approaches}

\par 
Metal defects initially relied on manual inspection, including Magnetic Particle Inspection (MPI), Ultrasonic Testing (UT), and Dye Penetrant Inspection (DPI)~\cite{lovejoy1993magnetic}. 
While these approaches have been fundamental in ensuring the quality of metals, they come with limitations such as inconsistency and potential human error.

\subsection{CNN}

\par
A compact Convolutional Neural Network was used alongside a cascaded autoencoder (CASAE) in the task of metal defects detection~\cite{Tao2018}. The compact CNN architecture aimed to classify defects, while the CASAE was used to localize and segment defects. The usage of CASAE resulted in accurate and consistent results even under complex lighting conditions or with irregular defects. The pipeline of the architecture started with passing the input image to the CASAE, which outputs a segmented image of the defects. The segments are then cropped and fed to the compact CNN to obtain classification results. Nevertheless, the architecture had limitations, as the input data must be manually labelled as segments, not as bounding boxes, which takes a lot of time and expense.

\subsection{RepVGG}

\par
The authors of~\cite{li2022} provides a reference for solving the problem of classifying aluminum profile surface defects. Defective images for training were obtained by means of digital image processing, such as rotation, flip, brightness, and contrast transformation. A novel model, RepVGG, with a convolutional block attention module (RepVGG-CBAM) was proposed. The model was used to classify ten types of aluminum profile surface defects.

\subsection{Faster R-CNN}

\par
Another novel approach proposes a method combining a classification model with an object recognition model for metal surface defects detection~\cite{wang2021}. An improved, faster R-CNN model is used to detect multi-scale defects better by adding spatial pyramid pooling (SPP)~\cite{mikolajczyk2017neural} and enhanced feature pyramid networks (FPN) modules~\cite{girshick2017feature}. The model aims to increase the detection accuracy of crazing defects by modifying the aspect ratio of the anchor. Non-maximum suppression is used to get the bounding box faster and better. Improved ResNet50-vd model is incorporated as the backbone of the classification model and object recognition model~\cite{he2019bag}. Detection accuracy and robustness are increased by adding the deformable revolution network (DCN)~\cite{zhu2018deformable}.

\subsection{YOLOv5}

\par
Among the deep learning models, the You Only Look Once (YOLO) algorithm stands out for its capabilities for object detection~\cite{redmon2016you}. YOLO reframes object detection as a single regression problem, predicting bounding boxes and class probabilities directly from full images in one evaluation. YOLOv5~\cite{bochkovskiy2020yolov4} is a famous version among the YOLO versions, as it was the first version with ultralytics support. YOLOv5 was used as a base model in metal defects detection ~\cite{Wang2022}, while adding a focus structure to the base network of the Darknet. Additionally, GIOU loss was chosen over L1 loss to focus on accuracy. The backbone feature extraction network of the YOLOv5 was retrained to improve the performance of the model. However, this model faced limitations due to its inability to detect small defects on metal surfaces.

\section{\uppercase{Datasets}}
\label{sec:datasets}

\par 
Several datasets have been created to provide research with standardized data for training and evaluating defects detection algorithms. This section focuses on three datasets - NEU-DET, GC10-DET and Multi-DET that were used for the work done in this research.

\subsection{Training Datasets}

\par 
GC10-DET is a dataset for surface defects collected in a real industry~\cite{gc10}. The dataset contains 2300 high-resolution images of surfaces with 10 different classes of defects, which are punching, weld line, crescent gap, water spot, oil spot, silk spot, inclusion, rolled pit, crease, and waist folding.

\par 
%%% WG
%%% The manuscript advised that the titles of the subsections and subsubsections should be titlecased: The heading of a subsection title must be with initial letters capitalized (titlecased).
%%% The Related Work section was completely rewritten.
NEU-DET is the Northeastern University (NEU) surface defect dataset~\cite{neu}. It contains 1800 grayscale images with 6 different classes of surface defects and 300 images per defect type. The defects' classes include rolled-in scale, patches, inclusion, scratches, crazing, and pitted surface.

\begin{figure}[!h]
    \centering
    \includegraphics[width=0.8\columnwidth]{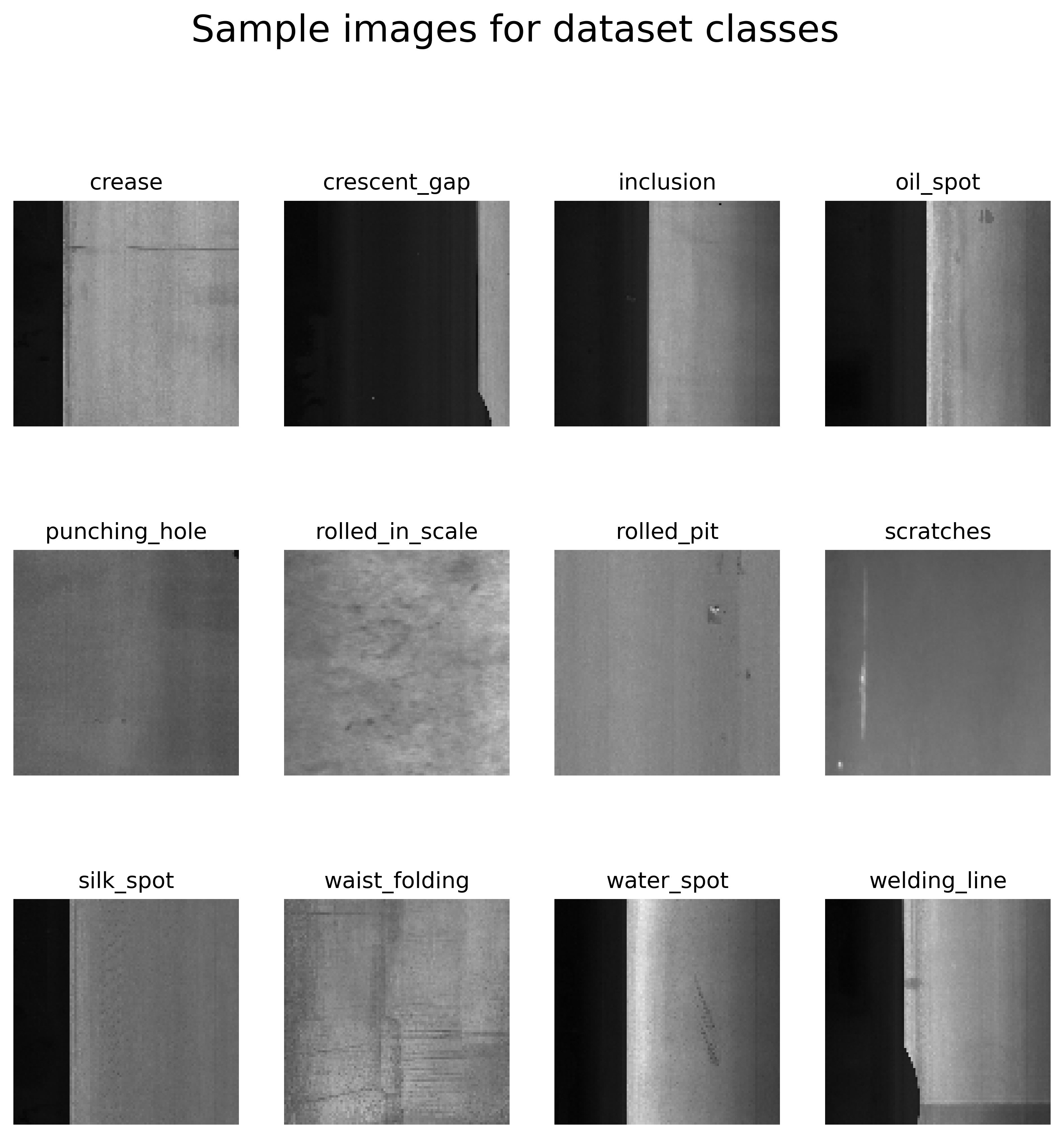}
    \caption{Dataset Defect Classes.}
    \captionsetup{justification=centering,singlelinecheck=false}
    \label{fig:defect Classes}
\end{figure}
\subsection{Multi-DET}
\par 
We introduce a new dataset called Multi-DET in order to address the limitations of the current datasets. The proposed dataset surpasses current datasets by offering increased diversity and density of defects per photo.

\par
\begin{figure}[!h]
  \centering
  \begin{tabular}{cc}
    \includegraphics[width = 2.75cm, height= 2cm]{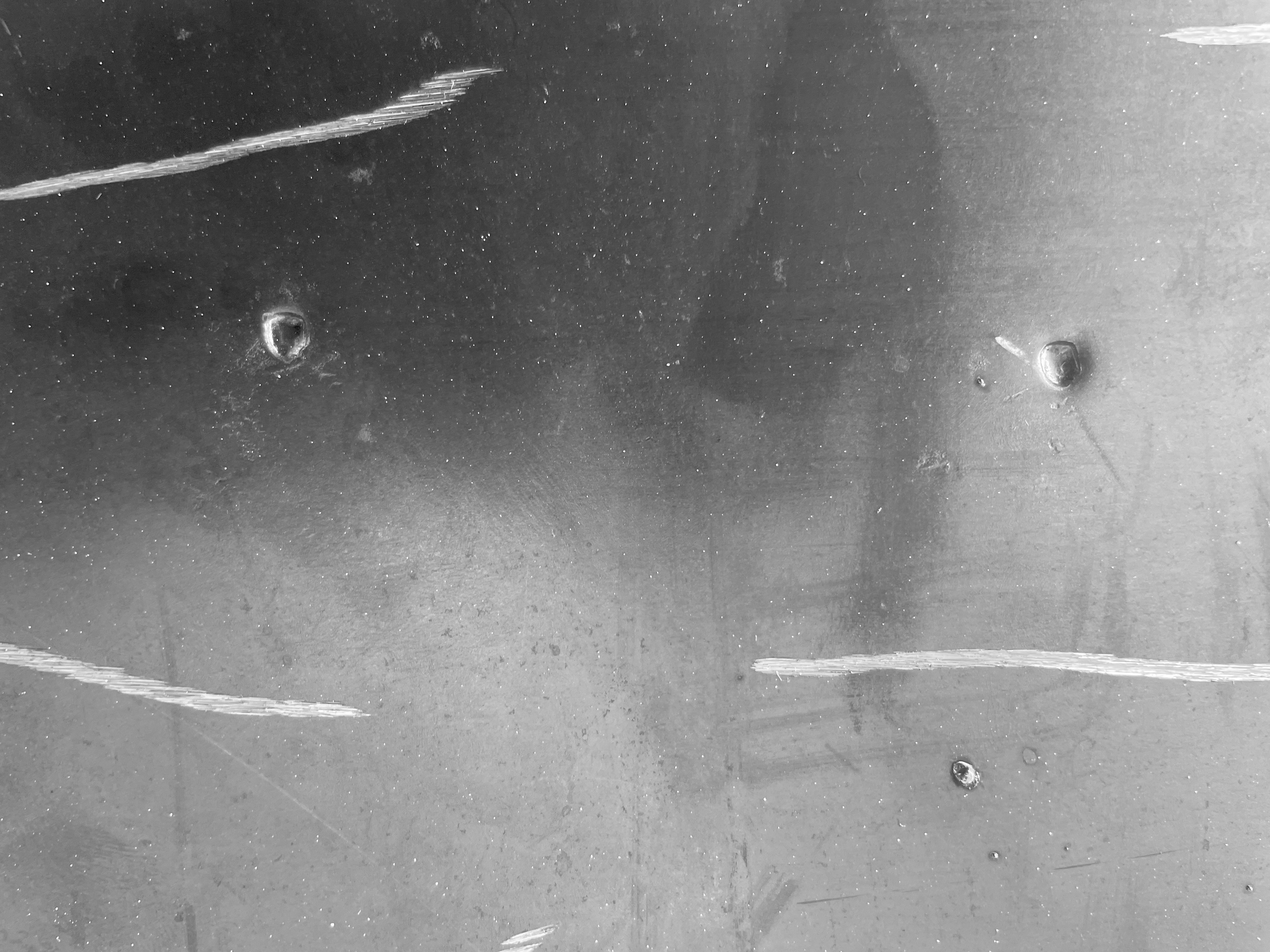}
    \includegraphics[width = 2.75cm, height= 2cm]{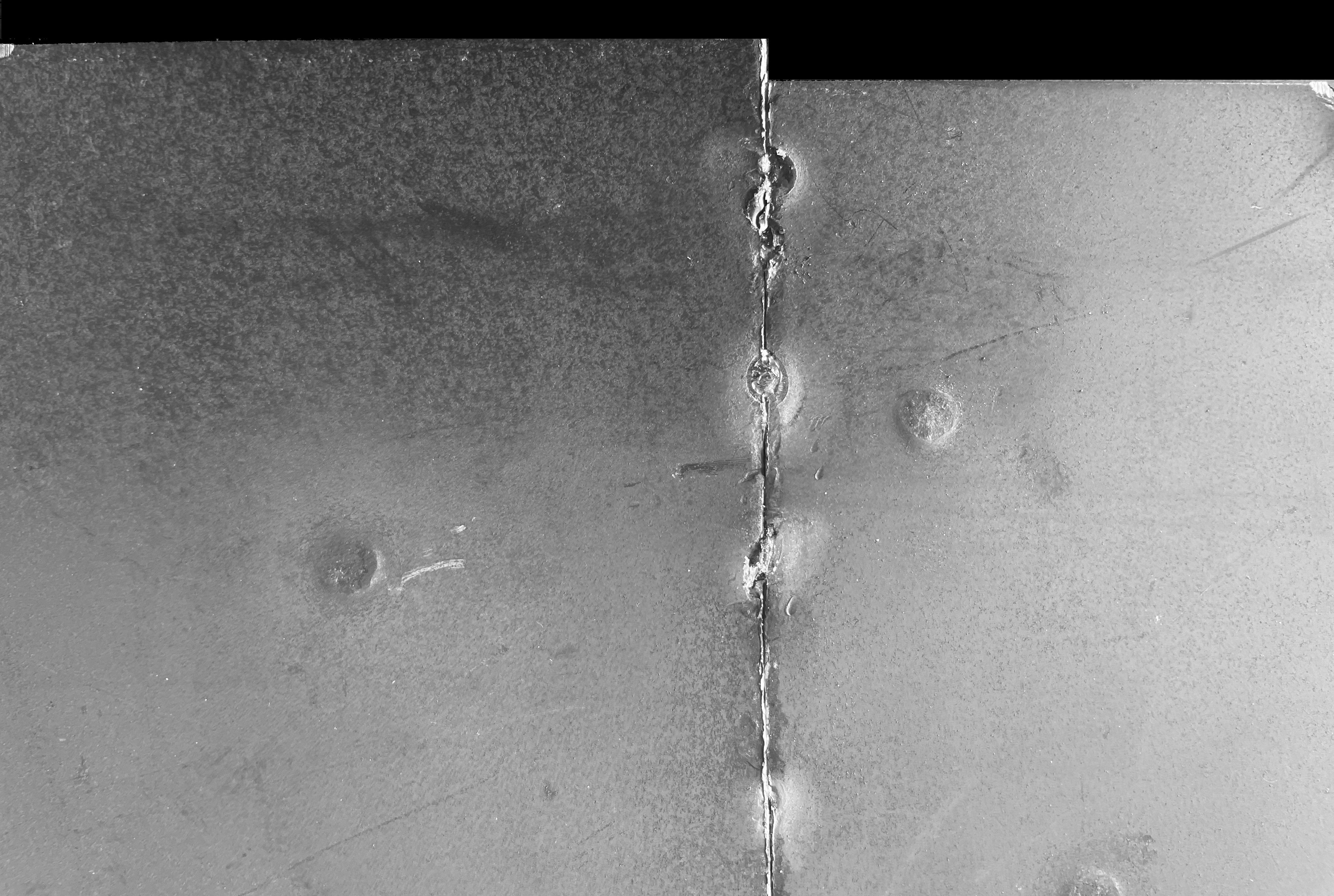} \\
    \includegraphics[width = 2.75cm, height = 2cm]{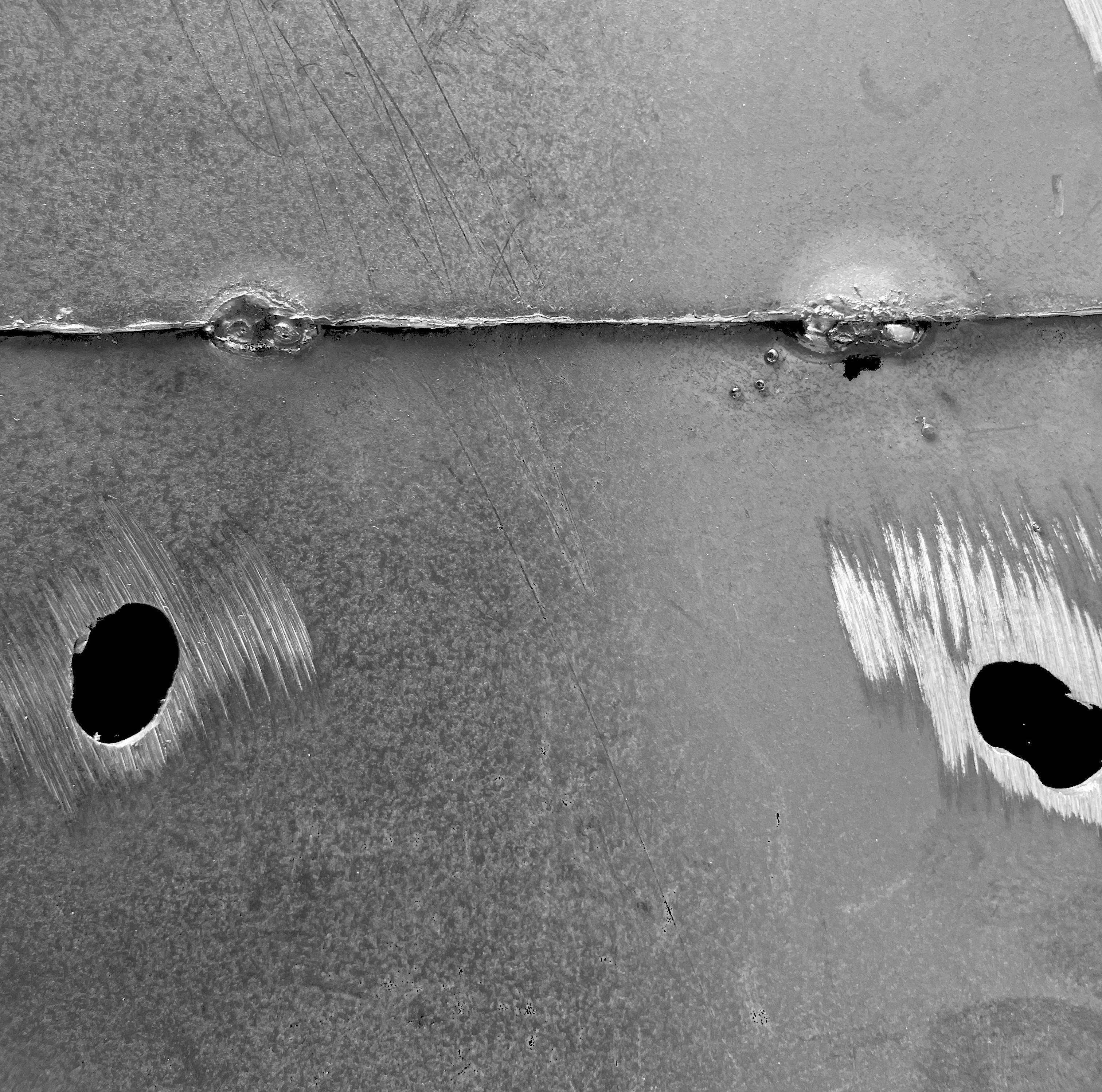} 
    \includegraphics[width = 2.75cm, height = 2cm]{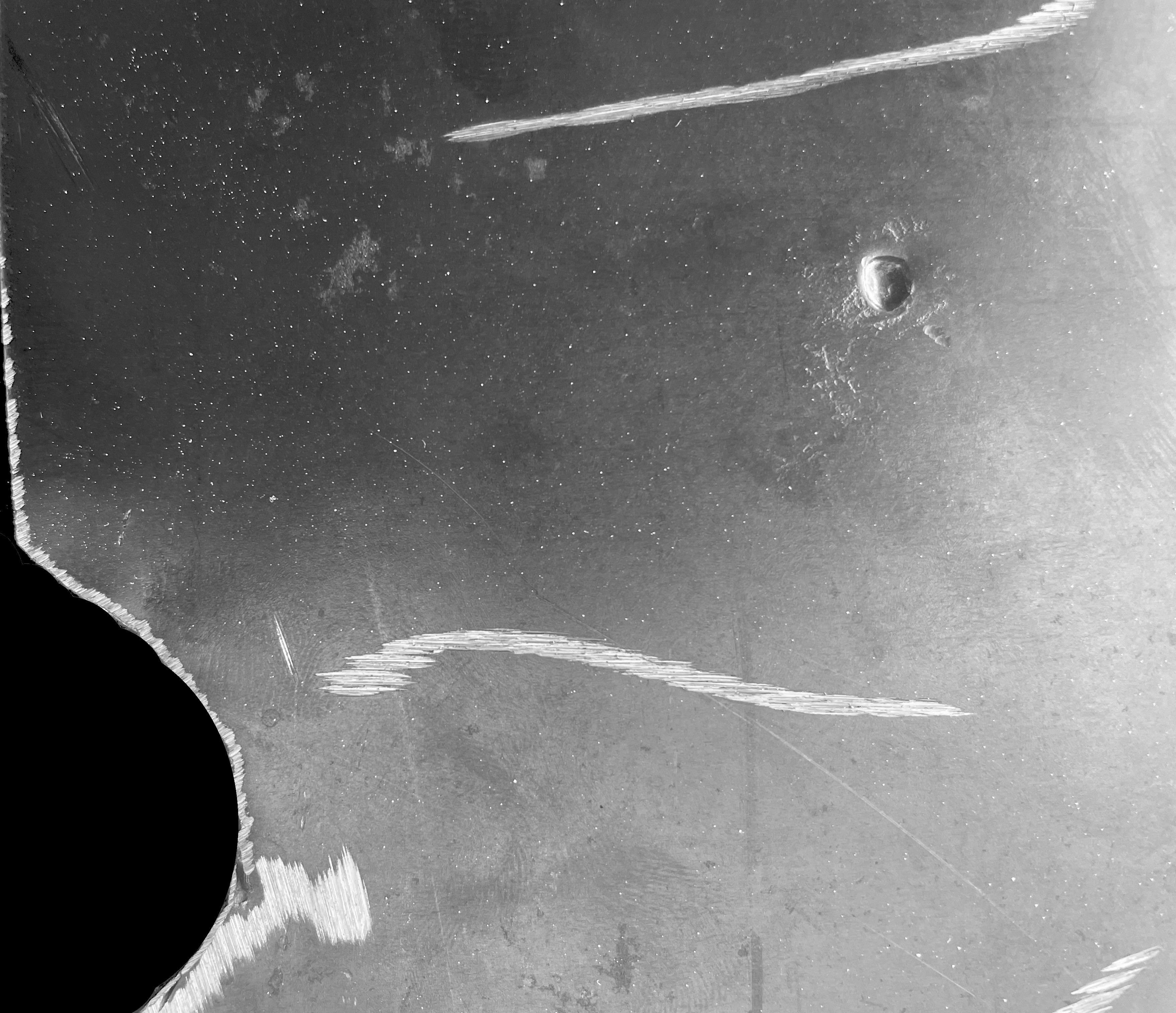} \\
  \end{tabular}
  \caption{Sample photos of Multi-DET dataset.}
  \label{fig:MULTI-DET}
\end{figure}

 
Multi-DET contains 300 high-resolution images for 8 classes. Unlike traditional datasets that feature repetitive defect types per image, our dataset covers a wide range of defects, including scratches, welding line, inclusion, water spot, oil spot, crescent gap , and variations in texture and color. Our approach mimics real-world conditions, where metal surfaces exhibit complex and overlapping defects. Figure \ref{fig:MULTI-DET} represents some samples of Multi-DET dataset.

\subsubsection{Dataset Collection}
\par 
The dataset collection process started with surface preparation, where metal samples were cleaned and smoothed to ensure a uniform pattern. Following this, different defects were introduced using various tools. Scratches were made using sharp instruments, welding lines were simulated using a welding machine, crescent gap were made using precise cutting tools, and inclusions were created using contaminant materials. Oil and water spots were applied using controlled droplets.

\subsubsection{Dataset Pre-processing}
\par 
The pre-processing of photos in Multi-DET is a crucial step in order to ensure quality and the uniformity in terms of resolution, lighting, and color balance. Pre-processing involves adjusting the brightness and contrast to compensate for any variations. In addition, image denoising is used to remove any unwanted noise that may interfere with the detection process. Image cropping is performed to focus on relevant parts of the metal, excluding information that does not contribute to the analysis. Images are converted to grayscale to enhance the feature extraction process. Data augmentation techniques are utilized to increase the variability of the dataset. This includes rotating, flipping, and scaling. These transformations represent the diversity of real-world conditions.

\subsubsection{Data Annotations}
\par
\begin{figure}[!h]
  \centering
   \includegraphics[width = 5.5cm]{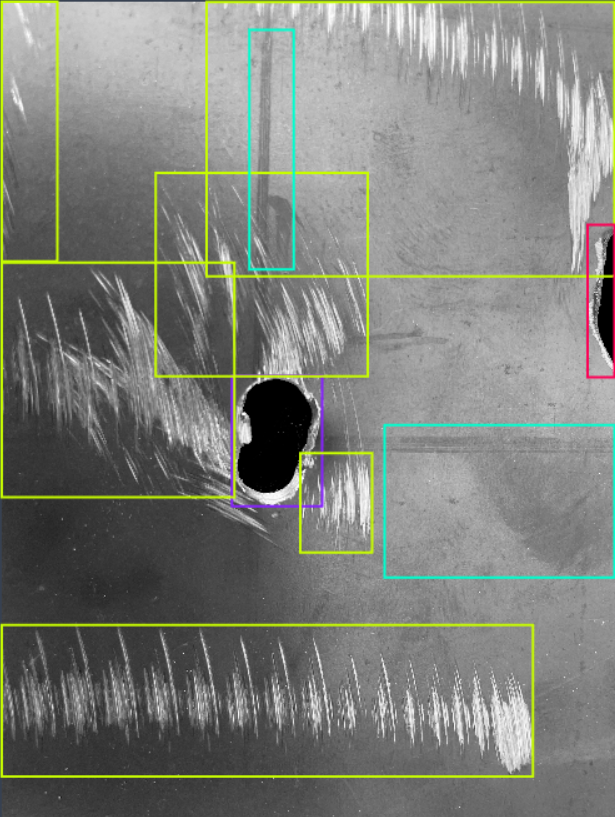}
  \caption{Defects bounding boxes annotation.}
  \label{fig:annotations}
 \end{figure}

 
For annotating Multi-DET dataset, Roboflow serves as the primary tool for creating the annotations for our dataset. Figure \ref{fig:annotations} illustrates the process of annotations.

\section{\uppercase{Experimental Work}}
\label{sec:experimenatl_work}
\subsection{Methodology}
\par 
This research presents an automated approach for detecting defects on metal surfaces utilizing the Vision Transformer (ViT) architecture. ViTs achieve enhanced accuracy via the self-attention mechanism inherent in transformer encoders. We utilize the ViT’s encoder as a feature extractor, which is then forwarded to a CNN, followed by a couple of Multi-Layer Perceptron (MLP) models for classification and localization. For the detection mechanism, we implement the anchor boxes method in order to dynamically detect any number of defects within the input image. This repository includes the source code for our implementation
\footnote{https://github.com/toqaalaa20/Metal-surface-defects-detection}. 

\subsection{Vision Transformers}

\par
\begin{figure}[h!]
    \centering
    \includegraphics[width=0.47\columnwidth]{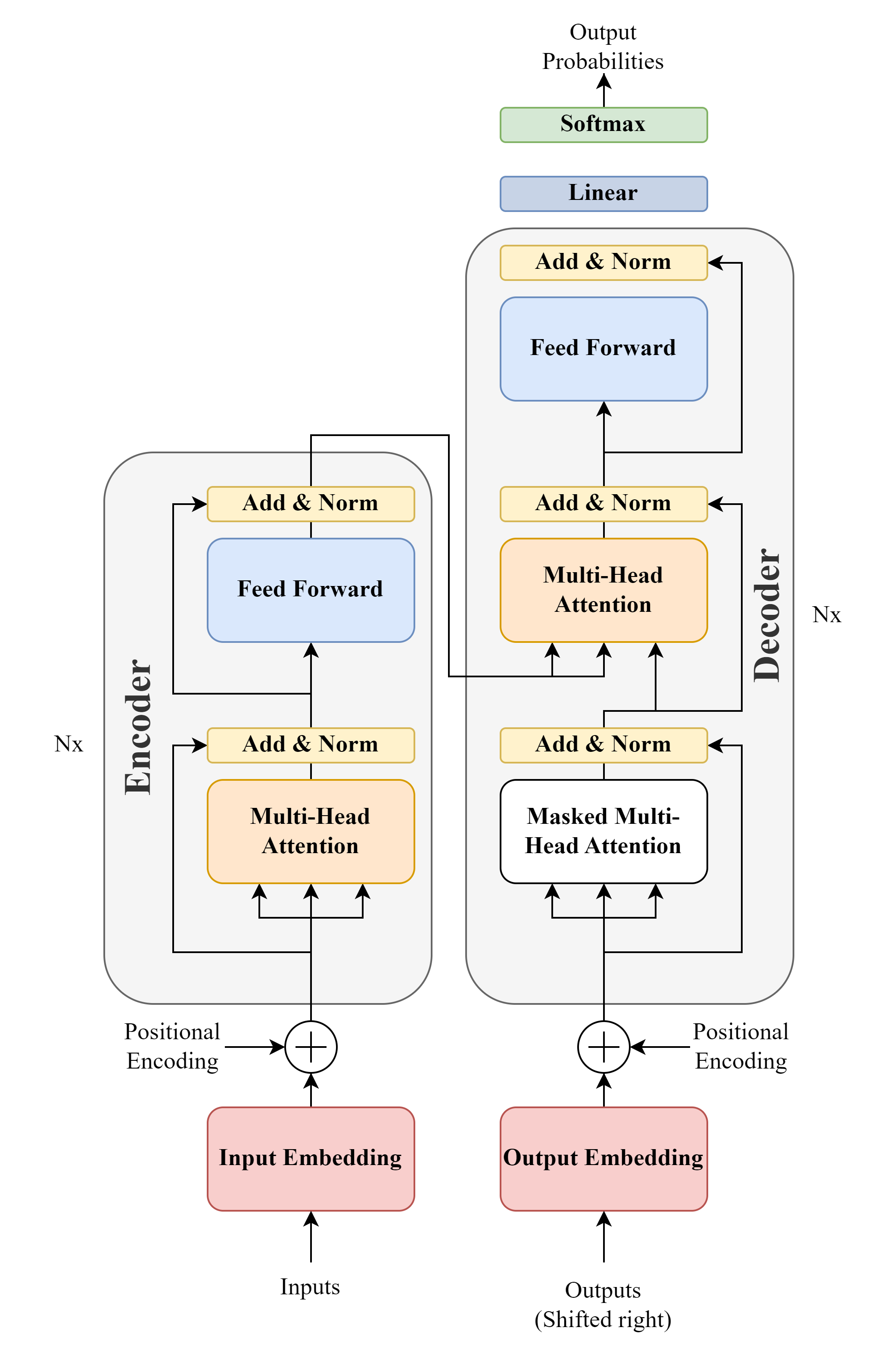}
    \caption{Transformer architecture.}
    \captionsetup{justification=centering,singlelinecheck=false}
    \label{fig:transformer}
\end{figure}

 
The architecture, Figure~\ref{fig:transformer}, of the ordinary transformer~\cite{NIPS2017_3f5ee243} was initially applied in Natural Language Processing (NLP). The main purpose of the transformer was to detect the relationships between the tokens through the self-attention mechanism. The architecture also added positional encoding to the input of the multi-head attention layers, which allowed the transformer to keep track of the order of the tokens. 

\par
\begin{figure}[h!]
    \centering
    \includegraphics[width=0.8\columnwidth]{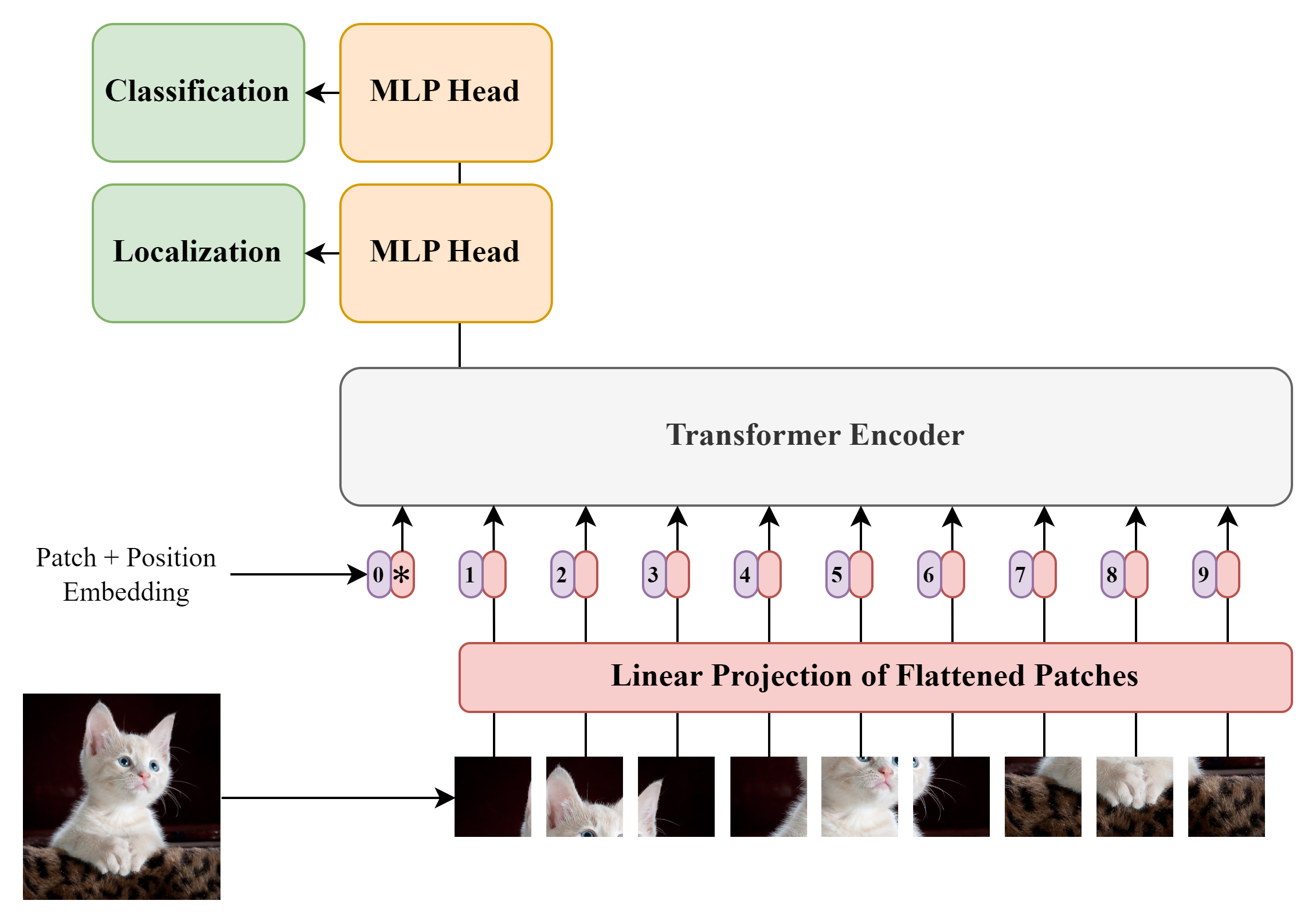}
    \caption{Vision Transformer Architecture.}
    \captionsetup{justification=centering,singlelinecheck=false}
    \label{fig:vit}
\end{figure}

 
The architecture in~Figure \ref{fig:vit} of the Vision Transformer (ViT) used the same basic structure of the transformer with the multi-head attention layers followed by the MLPs and the normalization layers~\cite{DBLP:journals/corr/abs-2010-11929}. However, the modification was on the input of the model, as the input was modified to take an image instead of a sequence of words. To use the same structure of the transformer, an input image is divided into patches and flattened. Positional embedding is added to the patches to keep track of the position of each patch. The encoded patches are fed to the ViT encoder. In order to perform any task of classification or localization, a learnable MLP is added to the output of the encoder.

\subsection{Anchor Boxes}
\par
\begin{figure}[!h]
    \centering
    \includegraphics[width=\columnwidth]{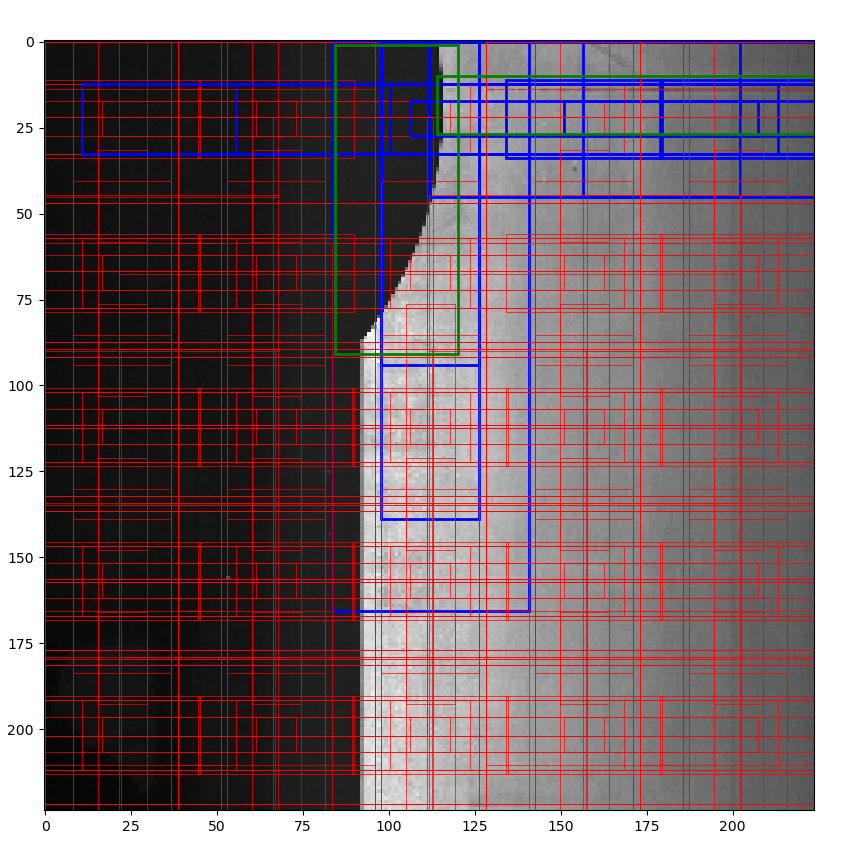}
    \caption{Initialized pre-defined anchor boxes. Red boxes are background boxes. Blue boxes are selected boxes before offset correction. Green boxes are the ground truth.}
    \captionsetup{justification=centering,singlelinecheck=false}
    \label{fig:anchor_boxes}
\end{figure}

 
In order to dynamically detect any number of defects in an image without being limited to a fixed number of defects per image, the anchor boxes method was used. The mechanism of anchor boxes was first introduced as a part of the You Only Look Once (YOLO) model architecture~\cite{redmon2016you}. At first, a set of pre-defined anchor boxes (Figure \ref{fig:anchor_boxes}) is defined. 

\par 
During training, each anchor box is assigned to a ground truth label depending on the Intersection over Union (IOU) value. If the IoU value is higher than an upper threshold, the anchor box gets assigned to a ground truth label. If the IoU is lower than a lower threshold, it is marked as background. If it is between the two thresholds, it is marked as discarded. 

\par 
After that, the offset in position from the ground truth is calculated for the assigned anchor boxes and set to zero for the background and discarded boxes. The class of the ground truth is also passed to the assigned anchor boxes. During prediction, a non-maximal suppression is applied on the predicted anchor boxes in order to choose the most suitable anchor box from the overlaying boxes with the same predicted class.

\subsection{Data Pre-processing}
\par 
In order to fit our model architecture, all images were resized and normalized to fit the input of the ViT encoder. Then, they were passed to an image processor, which prepares the images to be a suitable input for the ViT.

\par 
Offsets are calculated as the distance between the anchor box point and the ground truth point divided by the width or the height of the anchor box. This makes the offsets invariant to different scales of anchor boxes. Then, the offsets are passed on a min-max scaler to normalize them by subtracting the minimum value from the offset and dividing the result by the difference between the minimum value and the maximum value. All these normalizations were made to help the model predict the values without overshooting or overfitting.

\subsection{Model}
\par
\begin{figure*}[!h]
    \centering
    \includegraphics[width = \textwidth]{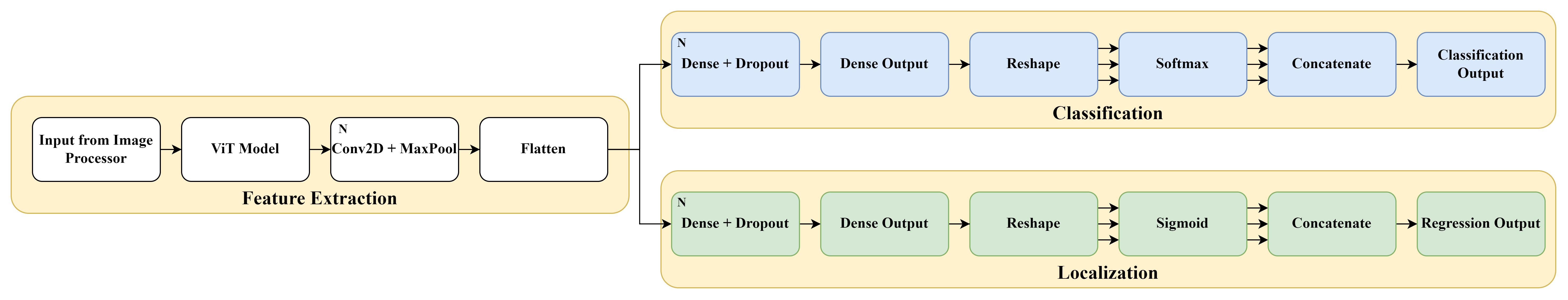}
    \caption{Our Model Architecture.}
    \captionsetup{justification=centering,singlelinecheck=false}
    \label{fig:model_architecture}
\end{figure*}

 
Our model architecture (Figure \ref{fig:model_architecture}) consists mainly of 4 parts: ViT model, CNN, Classification MLP, Regression MLP. 

\subsubsection{Feature Extraction}
\par 
After the image is pre-processed using the image processor, it is passed to the encoder of the ViT model. The output embeddings of the encoder are then passed on a CNN to process these embeddings and extract the features from them.

\par 
The output of the CNN is then flattened and shared to two different MLPs. The share mechanism is selected over using two different models for classification and localization to allow the model’s CNN to learn the common features between classification and localization. This will allow the model to understand the image features better and connect the two parts of detection together.
\subsubsection{Detection}
\par 
Each MLP consists of multiple dense and dropout layers followed by an output layer. The output layer is then reshaped to (number of anchor boxes, number of classes) in the classification MLP and reshaped to (number of anchor boxes, 4) in the regression MLP. This is done to apply Softmax in classification and sigmoid in regression individually on each sample (anchor box). Sigmoid is used as the values of the offsets are limited between 0 and 1 after scaling. This reshape will yield more meaningful results as applying Softmax on the whole number of samples will yield false results.

\par 
The output of the Softmax layers and sigmoid layers is then concatenated and reshaped to match the output shape, which is [(number of anchor boxes, number of classes), (number of anchor boxes, 4)].
\subsection{Loss Functions}
\par 
The model is compiled with modified versions of Categorical Cross-Entropy and Mean Square Error (MSE) for classification and regression, respectively.
\subsubsection{Classification Head}
\par 
In categorical cross-entropy, the modification aimed to handle the fact that most anchor boxes will be assigned to the background class, which will lead towards extreme bias towards the background class. In other words, it would be easier for the model to predict all classes as background than actually spotting features in the image. To eliminate this bias, the categorical cross-entropy is performed individually on each sample only if the true value is not a background. This is to mimic a two-level categorical cross-entropy, a lower level on each sample, and a higher level on the whole image with all samples. Algorithm \ref{alg:CCE} describes how the modified function works.
\SetKwComment{Comment}{/* }{ */}
\RestyleAlgo{ruled}
\begin{algorithm}[!h]
\caption{Modified Categorical Cross-entropy.}\label{alg:CCE}
\KwData{$y_{\text{true}},\ y_{\text{pred}}$}
\KwResult{$Categorical\ Cross\ Entropy\ Value$}

$N \gets len(y_{\text{true}})$\;
$M \gets len(y_{\text{true}}[0])$\;
$loss\gets 0$\;
\For{$i \gets 0$ \KwTo $N$}
{
    \For{$j \gets 0$ \KwTo $M$}
    {
        \If{$argmax(y_{\text{true}}[i][j]) \neq background$}
        {
            $loss \gets loss + \sum_{m=1}^{M} y_{\text{true}}[i][j][m] \log(y_{\text{pred}}[i][j][m])$\;
        }
    }
}
\KwRet{$loss$}
\end{algorithm}

\subsubsection{Localization Head}
\par 
In MSE, the modification aimed to handle the same problem that most offsets will be 0 due to being in a background class. The modification was to iterate over all samples and calculate the MSE individually for each sample only if the true values are not zeros, and then output the total MSE. Algorithm \ref{alg:MSE} describes how the modified function works.

\begin{algorithm}[!h]
\caption{Modified Mean Squared Error.}\label{alg:MSE}
\KwData{$y_{\text{true}},\ y_{\text{pred}}$}
\KwResult{$Mean\ Squared\ Error\ Value$}

$N\gets len(y_{\text{true}})$\;
$M\gets len(y_{\text{true}}[0])$\;
$loss\gets 0$\;
$count\gets 0$\;
\For{$i \gets 0$ \KwTo $N$}
{
    
    \If{$\sum_{m=1}^{M}(y_{\text{true}}[i][m]) \neq 0$}
    {
        $loss\gets loss + \sum_{m=1}^{M} ({y_{\text{true}}[i][m] -y_{\text{pred}}[i][m])}^2$\;
        $count\gets count + 1$\;
    }
}
$loss \gets \frac{loss}{count}$\;
\KwRet{$loss$}
\end{algorithm}

\section{\uppercase{Evaluation}}
\label{sec:evaluation}
\par 
This section discusses the evaluation metrics used to assess our model, the obtained results, the strengths, and the limitations of our approach.
\subsection{Metrics}
\par 
Evaluation metrics are essential for assessing the performance our model. In this study, we employ three key metrics: a modified version of accuracy for defect classification, a modified version of Mean Absolute Error for bounding box regression, and Mean Intersection over Union (Mean IOU) for bounding box localization.
\subsubsection{Accuracy: Defect Classification}
\par 
The modification on accuracy followed the same procedures mentioned in Algorithm \ref{alg:CCE}. The accuracy measures the proportion of the correctly classified defect instances out of the total instances.
\begin{equation}\label{accuracy}
\text{Accuracy} = \frac{\text{Number of correctly classified defects}}{\text{Total number of defects}}
\end{equation}

\subsubsection{Mean Absolute Error: Bounding Box Regression} 
\par 
The modification on MAE followed the same procedures mentioned in Algorithm \ref{alg:MSE}. MAE quantifies the average absolute distance between predicted bounding box coordinates  \( \hat{B}_i = (x_{\hat{B}_i}, y_{\hat{B}_i}, w_{\hat{B}_i}, h_{\hat{B}_i}) \), and ground truth bounding box coordinates \( B_i = (x_{B_i}, y_{B_i}, w_{B_i}, h_{B_i}) \) for each instance \( i \).

\begin{equation}
\begin{aligned}
\text{MAE} = \frac{1}{n} \sum_{i=1}^{n} \Big( & | x_{\hat{B}_i} - x_{B_i} | + | y_{\hat{B}_i} - y_{B_i} | \\
                                            & + | w_{\hat{B}_i} - w_{B_i} | + | h_{\hat{B}_i} - h_{B_i} | \Big)
\end{aligned}
\end{equation}
where \( n \) is the total number of instances.
\subsubsection{Mean IOU}
\par 
Mean IOU measures the spatial overlap between predicted and ground truth bounding boxes across all defect instances.
\begin{equation}
\text{IOU}(B_i, \hat{B}_i) = \frac{\text{Area of overlap}(B_i, \hat{B}_i)}{\text{Area of union}(B_i, \hat{B}_i)}
\end{equation}
where:
\begin{itemize}
    \item \(\text{Area of overlap}(B_i, \hat{B}_i)\) is the area where the predicted and ground truth bounding boxes overlap.
    \item \(\text{Area of union}(B_i, \hat{B}_i)\) is the area encompassed by both the predicted and ground truth bounding boxes.
\end{itemize}
Mean Intersection over Union (Mean IOU) is calculated as the average IOU across all bounding boxes:
\begin{equation}
\text{Mean IOU} = \frac{1}{n} \sum_{i=1}^{n} \text{IOU}(B_i, \hat{B}_i)
\end{equation}

\subsection{Results}
\par
\begin{figure}[!h]
  \centering
   \includegraphics[width = 0.8\columnwidth]{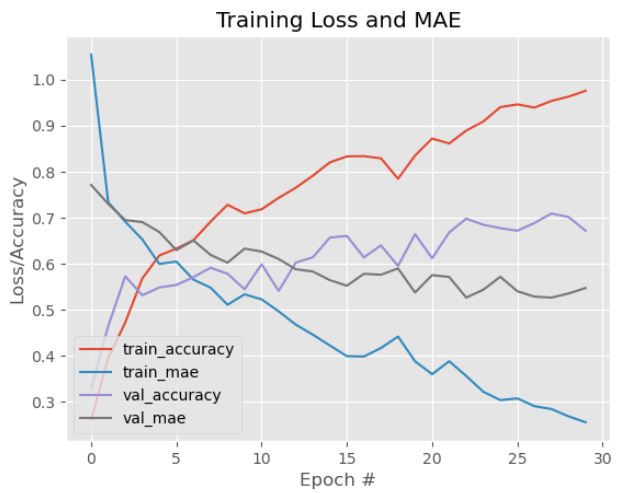}
  \caption{Evaluation metrics without using ViT.}
  \label{fig:cnn arch results}
 \end{figure}

\begin{figure}[!h]
  \centering
   \includegraphics[width = 0.8\columnwidth]{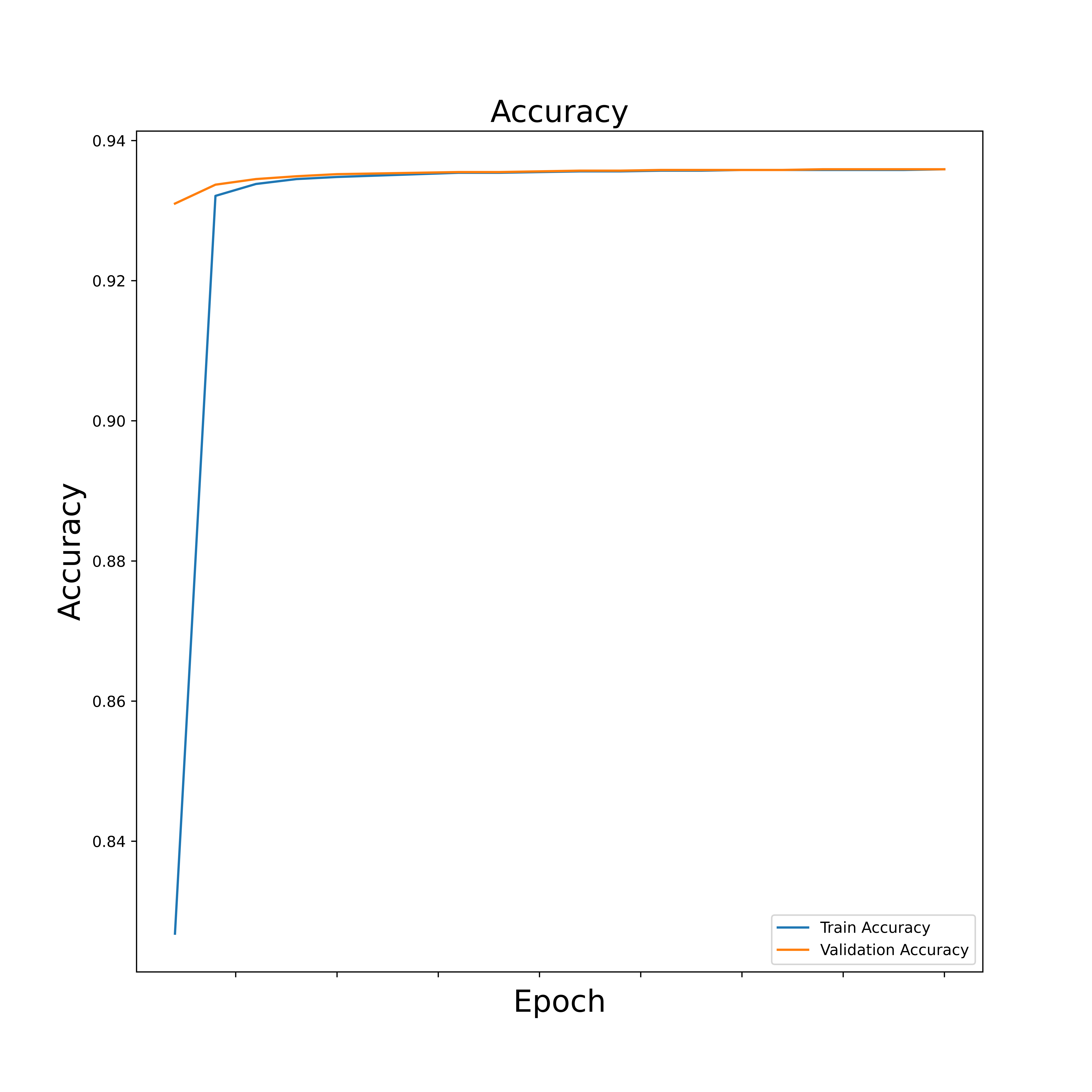}
  \caption{Model accuracy using ViT.}
  \label{fig:acc_vit}
 \end{figure}

\begin{figure}[!h]
  \centering
   \includegraphics[width = 0.8\columnwidth]{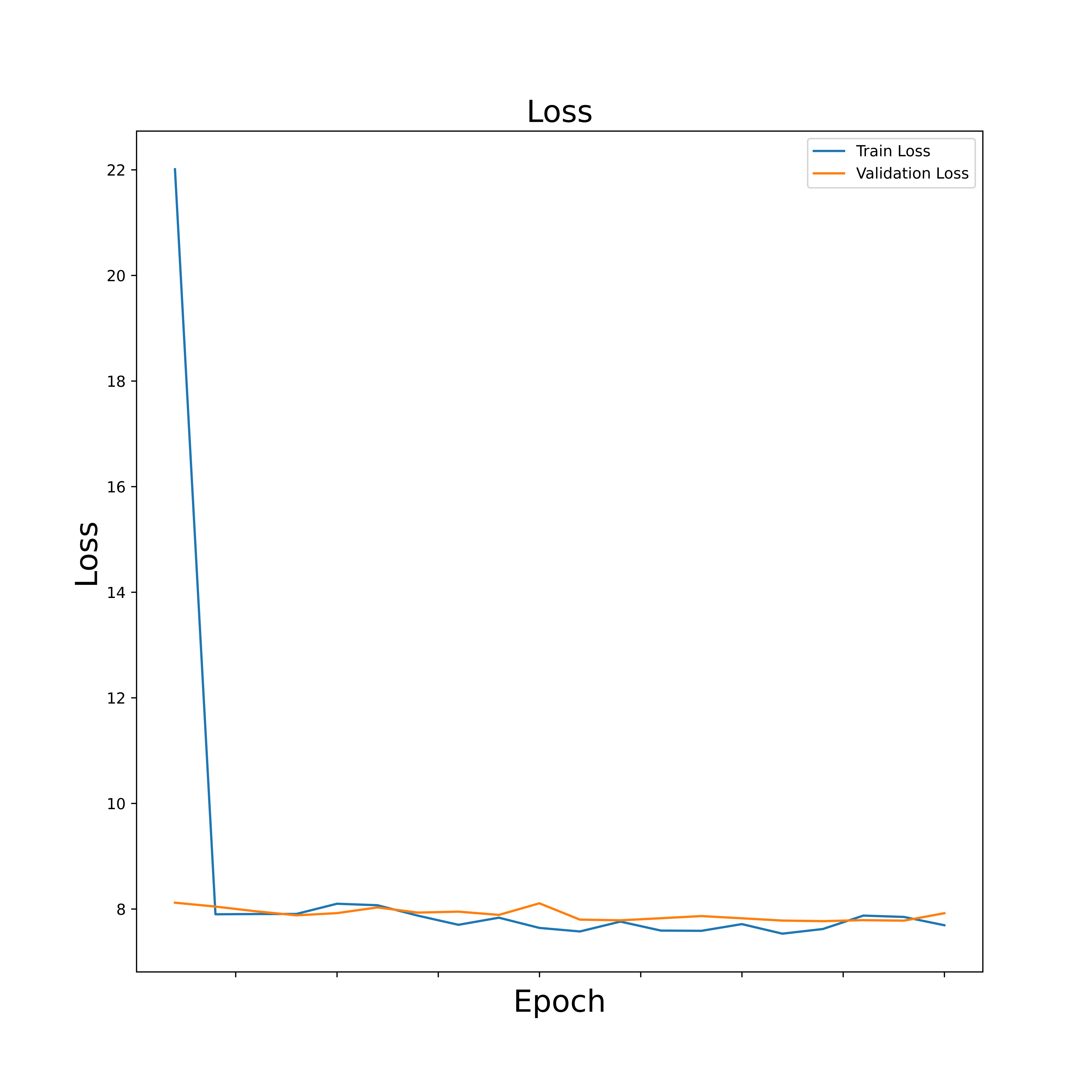}
  \caption{Model loss using ViT.}
  \label{fig:loss_vit}
 \end{figure}

\begin{figure}[!h]
  \centering
   \includegraphics[width = 0.8\columnwidth]{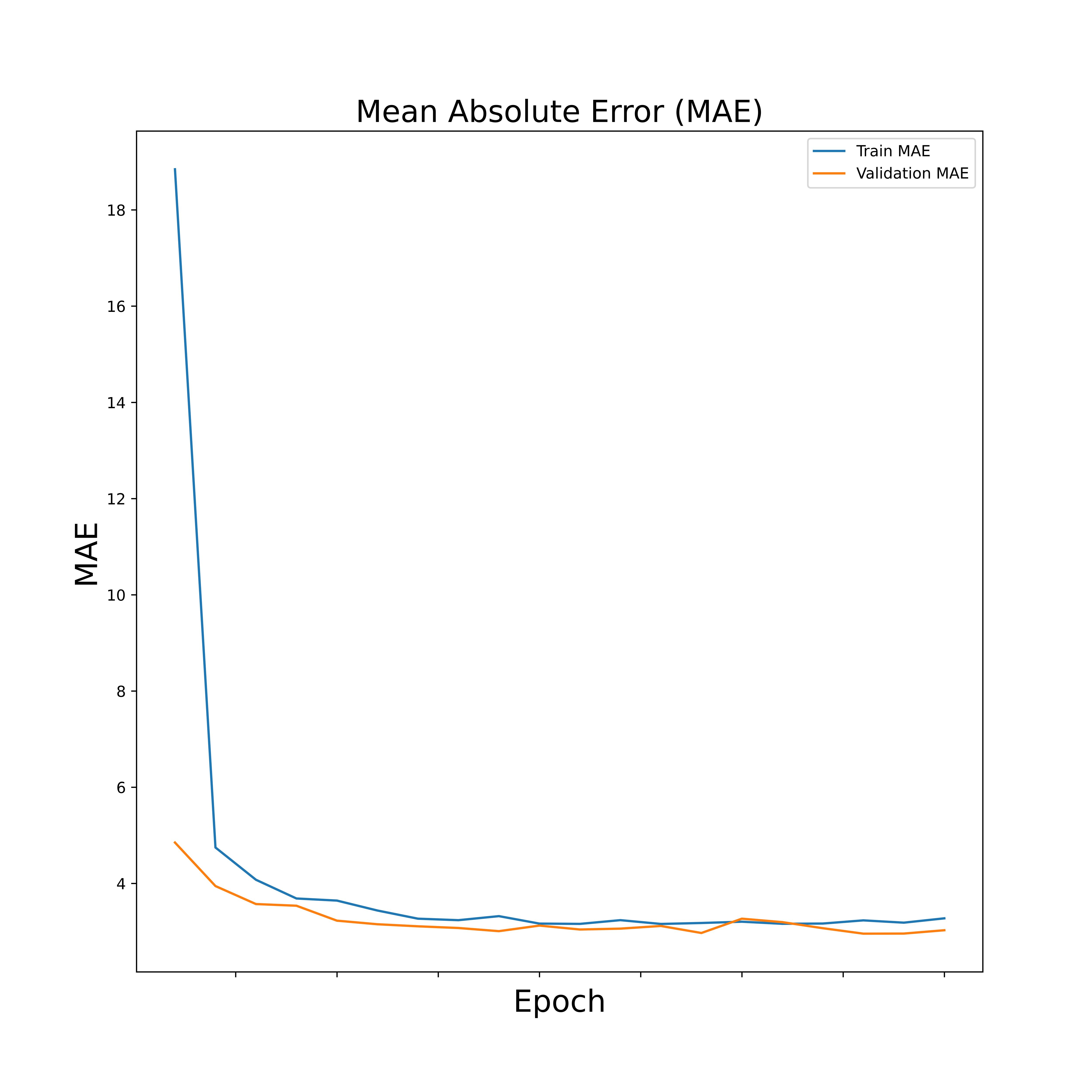}
  \caption{MAE using ViT.}
  \label{fig:mae_vit}
 \end{figure}

 
This section emphasizes the results of the loss functions, and the evaluation metrics illustrated earlier. The figures compare the results obtained by training our data on our model with and without using the ViT as our base feature extractor. Figure \ref{fig:cnn arch results} shows the training versus validation accuracy and MAE without using ViT as a feature extractor. As shown in the figure, the model is overfitting our data as there is a significant gap between the training and the validation results. Figure \ref{fig:acc_vit} shows the training and the validation accuracy using the ViT. Figure \ref{fig:loss_vit} shows the loss using the ViT. Figure \ref{fig:mae_vit} shows the MAE using ViT. As the figures show, using ViT as the feature extractor has addressed the problem of over-fitting and achieved high performance.

\subsection{Discussion}
\par 
In this study, we evaluated a metal surface detection algorithm using classification accuracy, MAE for bounding box regression, and Mean IOU for bounding box localization. Our findings indicate significant achievements and some areas for improvement.

\subsubsection{Performance Evaluation}
\par 
Our model has achieved a high accuracy of 93.5\% in classifying metal surface defects. This high accuracy emphasizes the effectiveness of our classification approach in detecting and distinguishing between different defects.

For bounding box regression, we calculated MAE to assess the accuracy of predicted bounding box coordinates compared to ground truth. The calculated MAE of \(3.2\) pixels suggests that our model can reasonably predict the dimensions and positions of the bounding boxes representing the defects.

Mean Intersection over Union (Mean IOU) was used to evaluate the spatial overlap between predicted and ground truth bounding boxes. Our model achieved a Mean IOU score of \(0.72\), indicating strong performance in accurately localizing metal defects within the bounding boxes. 

\subsubsection{Strengths and Limitations}
\par 
Our study demonstrates several strengths, including robust classification accuracy and effective bounding box localization capabilities. These strengths highlight the potential of our approach to contribute to automated quality control processes in metal manufacturing industries.

\par 
However, our approach also has limitations. For instance, the current model may struggle with detecting highly irregular defects due to limitations in the training data. In addition, the detection and the classification process are not fast enough due to the complex details of the transformer architecture. 

\par 
Addressing these limitations could involve adding more samples to Multi-DET dataset to introduce these variations and incorporating further development on the vision transformer to optimize the detection and classification processes.

\section{\uppercase{Conclusions}}
\label{sec:conclusion}
\par 
Automated defect detection on metal surfaces is a crucial research area as it contributes to various industries, like automotive and construction. Manual inspection methods are slow and subjective, calling for automated systems. This study proposes using Vision Transformers to overcome the limitations of traditional methods. ViTs, with their attention mechanisms, can capture complex defect patterns effectively. The research focuses on defect classification and localization, using pre-trained ViTs and transfer learning. By automating defect detection, the approach aims to improve product quality and reduce errors in metal manufacturing. The study addresses a research gap in applying ViTs to metal surface defect detection, contributing to the field. The promising results demonstrate accurate defect classification and precise defect localization. This research advances automated defect detection, benefiting multiple industries.

\par 
Our methodology offers a promising approach for addressing the challenges posed by metal defects in manufacturing and reshaping industries. However, there is still room for improvement, particularly in addressing the model's capability for detecting extremely overlapping and irregular shapes of defects. This can be done by adding degrees of freedom to the model while augmenting the training dataset. In addition, optimizing the model to work in real-time will levitate the model's performance. This limitation is due to the complexity of the ViT. Ultimately, this research paves the way for more effective defect detection, ensuring the production of high-quality metal products, and reducing operational challenges in various industries.

\section*{\uppercase{Acknowledgements}}

\par 
We would like to extend our sincere gratitude to Eng. Fatma Youssef, for her invaluable help and guidance throughout this project. Her expertise and thoughtful advice have played a crucial role in shaping the path and achievements of this research.

{\small
\bibliography{example}}
\end{document}